\documentclass[conference]{IEEEtran}
\IEEEoverridecommandlockouts
\usepackage{cite}
\usepackage{amsmath,amssymb,amsfonts}
\usepackage{algorithmic}
\usepackage{graphicx}
\usepackage{textcomp}
\usepackage{xcolor}
\usepackage{algorithm}
\usepackage{algorithmic}
\usepackage{booktabs}
\usepackage{tabularx}
\usepackage{float}
\usepackage{pifont}
\usepackage{epstopdf}
\def\BibTeX{{\rm B\kern-.05em{\sc i\kern-.025em b}\kern-.08em
    T\kern-.1667em\lower.7ex\hbox{E}\kern-.125emX}}
\begin{document}

\title{SSP-RACL: Classification of Noisy Fundus Images with Self-Supervised Pretraining and Robust Adaptive Credal Loss\\

\thanks{Key Laboratory of Medical Electronics and Digital Health of Zhejiang Province (MEDC202301) and National Student Entrepreneurship Practice Program of Fudan University (202310246001S)}
}

% \author{\IEEEauthorblockN{1\textsuperscript{st} Mengwen Ye}
% \IEEEauthorblockA{\textit{Academy for Engineering and Technology} \\
% \textit{Fudan University}\\
% Shanghai, China \\
% mwye22@m.fudan.edu.cn}
% \and
% \IEEEauthorblockN{2\textsuperscript{nd} Yingzi Huangfu}
% \IEEEauthorblockA{\textit{Academy for Engineering and Technology} \\
% \textit{Fudan University}\\
% Shanghai, China \\
% fyzhuang23@m.fudan.edu.cn}
% \and
% \IEEEauthorblockN{3\textsuperscript{rd} You Li}
% \IEEEauthorblockA{\textit{Academy for Engineering and Technology} \\
% \textit{Fudan University}\\
% Shanghai, China \\
% youli22@m.fudan.edu.cn}
% \and
% \IEEEauthorblockN{4\textsuperscript{th} Zekuan Yu*}
% \IEEEauthorblockA{\textit{Academy for Engineering and Technology} \\
% \textit{Fudan University}\\
% Shanghai, China \\
% yzk@fudan.edu.cn}
% }
\author{
\begin{minipage}[t]{0.45\textwidth}
\centering
1\textsuperscript{st} Mengwen Ye\\
\textit{Academy for Engineering and Technology} \\
\textit{Fudan University}\\
Shanghai, China \\
mwye22@m.fudan.edu.cn
\end{minipage}
\hfill
\begin{minipage}[t]{0.45\textwidth}
\centering
2\textsuperscript{nd} Yingzi Huangfu\\
\textit{Academy for Engineering and Technology} \\
\textit{Fudan University}\\
Shanghai, China \\
fyzhuang23@m.fudan.edu.cn
\end{minipage}
\\[1.5ex]
\begin{minipage}[t]{0.45\textwidth}
\centering
3\textsuperscript{rd} You Li\\
\textit{Academy for Engineering and Technology} \\
\textit{Fudan University}\\
Shanghai, China \\
youli22@m.fudan.edu.cn
\end{minipage}
\hfill
\begin{minipage}[t]{0.45\textwidth}
\centering
4\textsuperscript{th} Zekuan Yu*\\
\textit{Academy for Engineering and Technology} \\
\textit{Fudan University}\\
Shanghai, China \\
yzk@fudan.edu.cn
\end{minipage}
}

\newcommand{\pname}{Robust Data Ambiguation}
\newcommand{\pnames}{RDA}

% Self-defined macros
\newcommand{\swap}[3][-]{#3#1#2} % just an example

\newcommand{\R}{{\mathbb{R}}}
\newcommand{\N}{{\mathbb{N}}}
\newcommand{\Z}{{\mathbb{Z}}}
\newcommand{\Q}{{\mathbb{Q}}}
\newcommand{\E}{{\mathbb{E}}}
\renewcommand{\P}{{\mathbb{P}}}
\newcommand{\F}{{\mathcal{F}}}
\newcommand{\evalue}{\mathbb{E}}	
\newcommand{\svert}{\, \vert \, }
\newcommand{\cX}{\mathcal{X}}
\newcommand{\cY}{\mathcal{Y}}
\newcommand{\cH}{\mathcal{H}}
\newcommand{\cD}{\mathcal{D}_N}
\newcommand{\eu}{\operatorname{EU}}
\newcommand{\au}{\operatorname{AU}}
\newcommand{\tu}{\operatorname{U}}
\newcommand{\Prob}{P}
\newcommand{\prob}{p}
\newcommand{\argmin}{\operatorname*{argmin}}
\newcommand{\argmax}{\operatorname*{argmax}}
\newcommand{\on}[1]{\operatorname{#1}}
\newcommand{\dir}{\operatorname{Dir}} 
\newcommand{\fromto}{\longrightarrow}

\newcommand*{\defeq}{\mathrel{\vcenter{\baselineskip0.5ex \lineskiplimit0pt
			\hbox{\footnotesize.}\hbox{\footnotesize.}}}%
	=}
\newcommand{\hath}{\widehat{h}}
\newcommand{\hatp}{\widehat{p}}
\newcommand{\haty}{\widehat{y}}
\newcommand{\sety}{\widehat{Y}}

\newcommand{\db}{\set{M}}
\newcommand{\fkt}[1]{#1(\cdot)}
\newcommand{\chrfkt}[1]{\mathbb{I}_{#1}}
\newcommand{\kref}[1]{(\ref{#1})}
\newcommand{\convto}{\rightarrow}
\newcommand{\fft}[3]{#1 :  #2 \rightarrow #3}
\newcommand{\with}{\,  | \,}
\newcommand{\given}{\, | \,}
\newcommand{\sothat}{\, : \,}
\newcommand{\defi}{\stackrel{\on{df}}{=}}
\newcommand{\set}[1]{\mathcal{#1}}

\newcommand{\impl}{\Rightarrow}
\newcommand{\groesser}[1]{\raisebox{#1mm}{} \raisebox{-#1mm}{}}
\newcommand{\sgroesser}{\groesser{1.20}}
\newcommand{\xleftr}{\left( \groesser{1.35} }
\newcommand{\xleftg}{\left\{ \groesser{1.35} }
\newcommand{\fftm}[5]{\fft{#1}{#2}{#3} \, ,\, #4 \mapsto #5}
\newcommand{\gdw}{\Leftrightarrow}
\newcommand{\gdwbd}{\stackrel{\on{df}}{\Leftrightarrow}}
\newcommand{\est}{{est}}
\newcommand{\epd}{\Leftrightarrow_{\on{def}}}
\newcommand{\pref}{\succ}
\newcommand{\variance}{\mathbf{V}}

%%% MACROS
\newcommand{\textmacro}[2]{\newcommand{#1}{#2\xspace}}
\newcommand{\mathsymbol}[2]{\newcommand{#1}{\ensuremath{\mathit{#2}}\xspace}}
\renewcommand{\vec}[1]{\boldsymbol{#1}}
\newcommand{\cat}{\operatorname{Cat}}

\newcommand{\cmark}{\ding{51}}%
\newcommand{\xmark}{\ding{55}}%
\maketitle

\begin{abstract}
Fundus image classification is crucial in the computer aided diagnosis tasks, but label noise significantly impairs the performance of deep neural networks. To address this challenge, we propose a robust framework, Self-Supervised Pre-training with Robust Adaptive Credal Loss (SSP-RACL), for handling label noise in fundus image datasets. First, we use Masked Autoencoders (MAE) for pre-training to extract features, unaffected by label noise. Subsequently, RACL employ a superset learning framework, setting confidence thresholds and adaptive label relaxation parameter to construct possibility distributions and provide more reliable ground-truth estimates, thus effectively suppressing the memorization effect. Additionally, we introduce clinical knowledge-based asymmetric noise generation to simulate real-world noisy fundus image datasets. Experimental results demonstrate that our proposed method outperforms existing approaches in handling label noise, showing superior performance. 
\end{abstract}


\section{Introduction}
Fundus image classification is pivotal in diagnosing various ocular diseases \cite{RN705}. Researchers need a large amount of labeled fundus images to improve the classification performance of deep neural networks. Given the limited availability and high costs of annotated medical image data, self-supervised learning has demonstrated significant maturity in the field of fundus image classification by utilizing unlabeled data to enhance network generalization capabilities. Yet, fundus image annotation quality can be compromised by expertise, demanding years of specialized training and domain knowledge. Mislabeling is also likely due to the high feature similarity and image quality variations. Consequently, fundus image datasets often inevitably contain label noise \cite{RN706}. Fundus images with noisy labels can confuse supervised training, resulting in the model overfitting noisy labels and severely reducing classification and generalization performance \cite{RN728}. Given the high demands for computer-aided diagnosis of ocular disease, addressing the issue of noisy labels is crucial. 

In recent years, researchers have proposed various methods to address the problem of noisy labels in natural images datasets, including regularization methods \cite{zhou2021learning,liu2020early,lu2021confidence}, loss function design \cite{zhang2018generalized,cheng2020learning,feng2021can}, sample selection strategies \cite{xia2023combating,patel2023adaptive,RN728}, label correction techniques \cite{tanaka2018joint,lu2022selc,wu2021learning}, self-supervised learning \cite{cheng2021demystifying} and integrated approaches \cite{li2023disc,sheng2024adaptive,albert2023your}. Fundus datasets often contain hard samples and are imbalanced, so these methods for natural images can not achieve good performance on fundus datasets. Classification on highly noisy, imbalanced fundus image datasets is still a challenging task. Despite its clinical importance, research on addressing label noise in fundus images remains limited. Several methods have been proposed to tackle this issue. For example,\cite{ju2022improving} proposed an uncertainty estimation-based framework to handle disagreement label noise from inconsistent expert opinions and single-target label noise from biased aggregation of individual annotations. \cite{hu2023fundus} addressed label noise by combining data cleaning, adaptive negative learning, and sharpness-aware minimization.\cite{galdran2020cost} employed a method involving cost-sensitive classification constraints and atomic sub-task modeling, resulting in a 3-5$\%$ improvement in quadratic-weighted kappa scores at a negligible computational cost. However, these methods are often complex and require additional precisely annotated set.
\begin{figure*}[htbp]
\centering
\includegraphics[width=\textwidth]{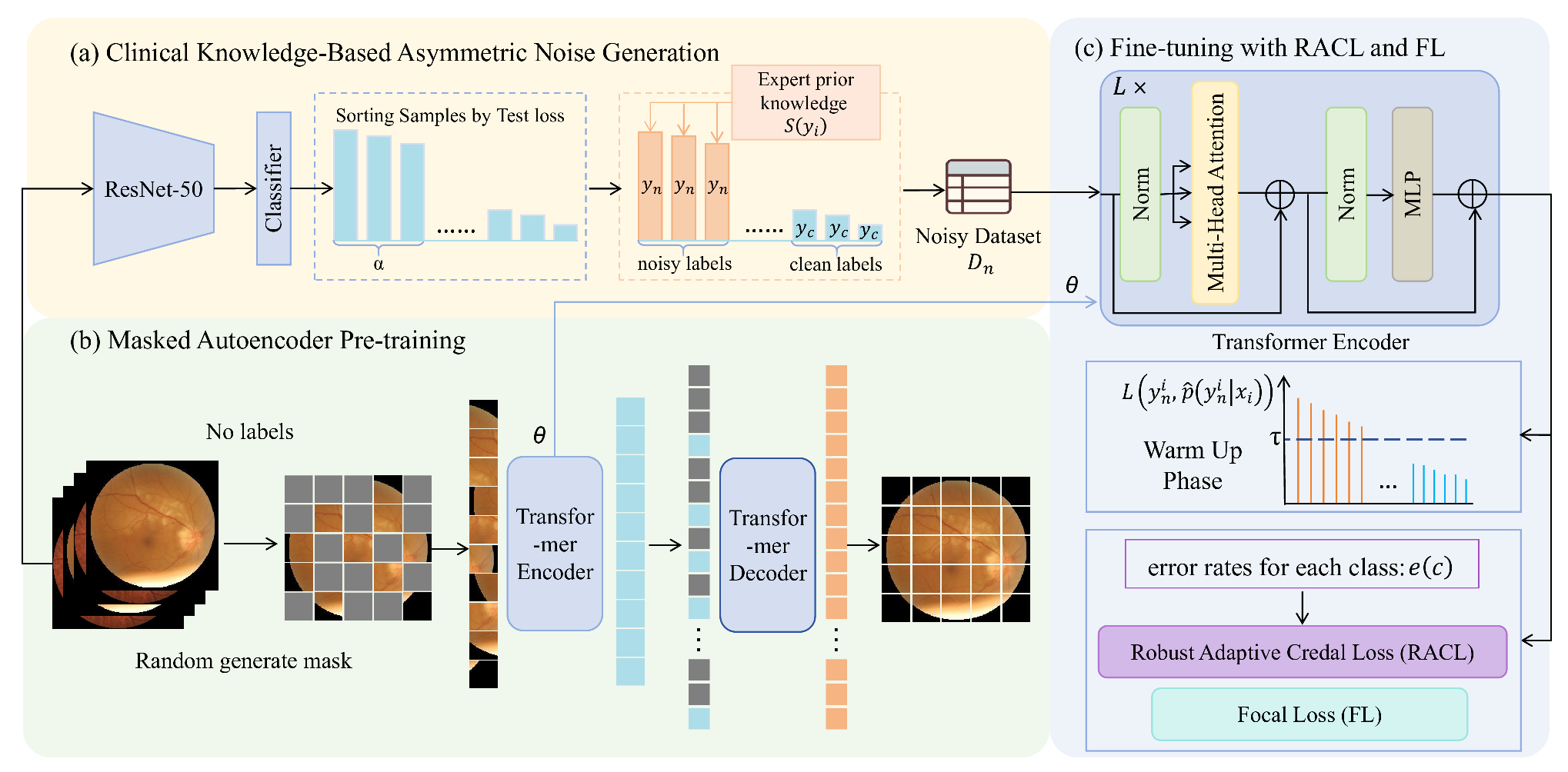}
\caption{The structure of the overall framework}
\label{fig}
\end{figure*}

To bridge this gap, we propose a novel method, Self-Supervised Pre-training and Robust Adaptive Credal Loss (SSP-RACL), for robust classification of noisy fundus images. Self-supervised pre-training (SSP) is robust against label noise and encompasses four strategies: innate relationship, generative, contrastive, and self-prediction\cite{huang2023self}. We adopt for the Masked Autoencoder (MAE) approach, classified under the generative strategy, due to its notable scalability. Our approach is simple but effective, and does not rely on additional evaluation set. More importantly, it significantly enhances the ability to handle noisy data, ensuring the reliability of the model in clinical applications.
\section{Methodology}
Initially, we employed clinical knowledge-based asymmetric noise generation techniques to create a simulated real-world noisy dataset. Subsequently, we employed MAE on this noisy dataset to ensure that the encoder captures fundamental features unaffected by label noise. Following this, we fine-tuned the model by integrating Robust Adaptive Credal Loss (RACL) with Focal Loss (FL). We innovatively propose an adaptive adjustment of the label relaxation parameter $\alpha$, offering increased flexibility in handling label noise. Additionally, the incorporation of FL further emphasizes focus on hard-to-classify samples. Fig.\ref{fig} illustrates the structure of the overall framework. 
\subsection{Clinical Knowledge-Based Asymmetric Noise Generation}
In studies of label noise in natural images, datasets are typically configured to symmetric noise and pairflip noise. However, fundus images annotated by physicians often contain systematic errors, which are mostly asymmetric. For example, in diabetic retinopathy grading tasks, the similar appearance of adjacent grades makes them more prone to misdiagnosis. To simulate these specific misdiagnosis patterns known in clinical practice, we propose clinical knowledge-based Asymmetric Noise Generation. Firstly, 30$\%$ of the samples in the dataset are randomly selected according to the original category proportions to form the training set. This subset is used to train a pre-trained ResNet-50 model. Next, the samples in test set are evaluated by the cross-entropy loss function. We select samples with the highest loss at a proportion \(r_n\) as candidates for noisy samples as high loss typically indicates low confidence in model predictions. These samples are likely misdiagnosed due to poor image quality, unclear features, or incorrect original labels. 
When generating noisy labels, instead of randomly assigning a different class to each selected sample, labels are adjusted based on expert knowledge. The noisy dataset \(D_n\) is defined as follows:
\begin{equation}
D_n = \{(x_i, y_i)\}_{i=1}^N\label{eq}
\end{equation}
where \(x_i\) represents the sample, and \(y_i\) is the corresponding label, defined by:
\begin{equation}
y_i = 
\begin{cases} 
y_c & 1 - r_n \\
y_n \text{ randomly selected from } S(y_i) & r_n 
\end{cases}\label{eq}
\end{equation}
where \(S(y_i)\) is the set of potential misdiagnosis labels determined based on expert knowledge, \(y_n\) represents the noisy label randomly selected from this set, \(y_c\) is the clean label, and \(r_n\) is the noise rate, indicating the probability that each sample is assigned a noisy label.
\subsection{Masked Autoencoder Pre-training}
Recent studies have demonstrated the effectiveness of MAE in pre-training for natural image analysis \cite{he2022masked}. By reconstructing complete images from partially masked inputs, the Vision Transformer encoder aggregates contextual information to infer the masked regions of the image. This capability to integrate contextual details is particularly crucial in fundus image analysis, as different anatomical structures are interdependent and closely connected. Therefore, we employ MAE to improve fundus image classification tasks with noisy labels. The fundus images are divided into non-overlapping patches, and around 80$\%$ of patches are randomly masked to reduce redundant information and encourage the model to learn global features. For each unmasked patch, a token is generated through linear projection and the addition of positional embeddings. Subsequently, the tokens undergo a shuffle operation to randomize their order, and then the top 20$\%$ are selected. These selected tokens serve as input for the Transformer encoder, as depicted in Fig. 1.  These feature vectors are then concatenated with vectors from positional information of masked patches. The decoder receives these concatenated vectors as input, and the output of the decoder is transformed back to the pixel scale through a linear layer. 
\subsection{Robust Adaptive Credal Loss}
Algorithm 1 details the steps of the RACL algorithm, which is used to handle noisy labels in downstream classification tasks and inspired by \cite{lienen2023conformal,wang2024creinnscredalsetintervalneural,lienen2024mitigating,caprio2024credallearningtheory}.
In traditional probabilistic learning settings, deterministic target labels $y_i \in \cY$ are transformed into degenerate probability distributions $p_{y_i} \in \P(\cY)$:
\begin{equation}
p_{y_i} = 
\begin{cases} 
1 & \text{if } y = y_i \\
0 & \text{otherwise}
\end{cases}\label{eq}
\end{equation}

This representation implies full plausibility for the observed training label $y_i$, while regarding other labels as fully implausible. However, traditional methods completely rely on individual label information, making them overly sensitive to noisy labels and consequently degrading model performance. The credal set is an effect approach to address this issue. Representing the target distribution as a set of possible distributions, rather than a single point estimate, credal sets account for the uncertainty associated with each training instance. This approach enables the model to consider a range of potential true distributions, accommodating scenarios where the observed label may be incorrect. Rather than treating a mislabeled instance as a definitive representation of the true class, credal sets allow the model to consider alternative labels that may better align with the underlying data distribution, enhancing the model's robustness.
Based on possibility theory \cite{dubois2001possibility}, we define the possibility distribution $\pi_i(y)$ for $x_i$ via confidence threshold. This assigns a probability to each class in $\cY$, considering them as candidates for the true outcomes associated with $\vec{x}_i$. $\pi_i(y)$ can be interpreted as an upper bound on the true probability $p^*(y \given \vec{x}_i)$ given $\vec{x}_i$. Specifically, $\pi_i(y)$ is set to 1 when $y$ equals the observed label $y_i$ or the model's predicted probability $\hatp(y \, | \, \vec{x}_i)$ exceeds the threshold $\beta$, thereby considering this label as a fully credible candidate. For all other cases, $\pi_i(y)$ is set to $\alpha$, indicating that even for labels with predicted probabilities below $\beta$, their possibility is not completely ruled out but instead assigned a lower possibility value. The mathematical representation is as follows:
\begin{equation} 
    \pi_i(y) = \begin{cases} 
        1 & \text{if } y = y_i \text{ or } \hat{p}(y|x) \geq \beta \\
        \alpha(c) & \text{otherwise}
    \end{cases}\label{eq}
\end{equation}
where $\beta, \alpha(c) \in [0,1]$. The threshold \(\beta\) is crucial for determining the plausibility of labels. As the model is relatively uncertain in the first epochs, one should not spent too much attention to the predictions. As training progresses and the model becomes more confident, smaller $\beta$ values can be used to incorporate more candidate labels in the confidence-driven possibility elicitation process. Specifically, \(\beta\) is updated according to the following decay schedule:
\begin{equation}
    \beta_T = \beta_1 + \frac{1}{2} (\beta_0 - \beta_1) \left( 1 + \cos\left( \frac{T}{T_{\text{max}}} \pi \right) \right)
\end{equation}
where \( T \) and \( T_{\text{max}} \) denote the current and maximum number of training epochs, respectively, while \( \beta_0 \) and \( \beta_1 \) represent the start and end values for \(\beta\). This dynamic adjustment ensures that the model remains cautious during the early training stages and progressively adjusts to more confident predictions as training continues \cite{lienen2024mitigating}. $\alpha(c)$ is a label relaxation parameter that is dynamically tuned based on the error rates $e(c)$ of each class. We capitalize on the characteristic that clean samples have lower losses while noisy samples result in higher losses during the warm-up phase of model training. To calculate $e(c)$ for each class, we use cross-entropy loss to assess the loss for each training sample during this warm-up phase. The error rate for a class is then defined as the proportion of samples from that class whose loss exceeds a threshold $\tau$:
\begin{equation} 
e(c) = \frac{|\{x \in \mathcal{D}_c : L(y, \hat{p}(y|x)) > \tau\}|}{|\mathcal{D}_c|}
\label{eq}
\end{equation}
where \(\mathcal{D}_c\) is the set of samples belonging to class \(c\). Then update \(\alpha(c)\) for each class based on its error rate:
\begin{equation}
    \alpha(c) = \frac{k}{e(c)+\epsilon}
\label{eq}
\end{equation}
where \(k\) and \(\epsilon\) are constants chosen to scale and stabilize the adjustments, ensuring \(\alpha(c)\) remains within [0, 1).
We can derive the credal targets $Q_\pi$ induced by $\pi_i(y)$, which define the set of all possible probability distributions $p$ that satisfy the condition: for any subset $Y \subseteq \mathcal{Y}$, the sum of probabilities for all events within 
$Y$  does not exceed the maximum possibility value corresponding to the events in $\pi_i$ \cite{lienen2023conformal}:
\begin{equation} 
Q_{\pi_i} \defeq \Big\{ p \in \mathbb{P}(\mathcal{Y}) \, \vert \, \forall \, Y \subseteq \mathcal{Y} : \, \sum_{y \in Y} p(y) \leq \max_{y \in Y} \pi_i(y) \Big\}\label{eq}
\end{equation}
Probability distribution $p^r$ that can be obtained by projecting the predicted probability $\hat{p}$ onto the boundary of the credal set $Q_{\pi_i}$. This projection adjusts $\hat{p}$ to ensure it complies with the constraints defined by the possibility distribution $\pi_i$, which specifies the upper bounds of probabilities for each class. $p^r(c)$ is given as follows:
\begin{equation}
    p^r(y) = \begin{cases}
(1 - \alpha(c)) \cdot \frac{ \hatp(y)}{\sum_{y' \in \mathcal{Y} : \pi_i(y')=1} \hatp(y')} & \text{if } \pi_i(y) = 1 \\
\alpha(c) \cdot \frac{ \hatp(y)}{\sum_{y' \in \mathcal{Y} : \pi_i(y')=\alpha(c)} \hatp(y')} & \text{otherwise}
\end{cases}\label{eq}
\end{equation}

For the dataset $\{ (\vec{x}_i, \mathcal{Q}_{\pi_i}) \}$ based on $\mathcal{Q}_{\pi_i}$ 
, we train the model by minimizing the Optimistic Superset Loss (OSL) \cite{hullermeier2014learning}. The core of OSL lies in selecting the probability distribution $p$ from the credal set $Q_\pi$ that minimizes the loss function $\mathcal{L}(p, \hatp)$. Specifically, it involves calculating the loss associated with all potential distributions $p$ within 
$\mathcal{Q}_{\pi}$ and selecting the smallest of these losses for optimization. Furthermore, the computational complexity of the OSL is equivalent to common loss functions, and the function is convex, which means that the associated optimization problem can be effectively solved, guaranteeing the identification of a global optimum \cite{lienen2021label}. The loss is given as follows:
\begin{equation}
    \mathcal{L}^*(\mathcal{Q}_{\pi} , \hatp) = \begin{cases}
        0  & \text{if } \hatp \in \mathcal{Q}_{\pi} \\
        D_{KL}(p^r \, || \, \hatp) & \text{otherwise} 
    \end{cases} \, ,
    \label{eq}
\end{equation}
where Kullback-Leibler Divergence $D_{KL}$ is used to measure the difference between two probability distributions. Through RACL, we can effectively reduce the model's reliance on incorrect training labels. Additionally, it enables us to be more cautious during prediction, avoiding erroneous judgments about potential mislabeled data. Therefore, this approach decreases the dependency on learning from noisy labels, while ensuring that the model's performance with clean samples remains unaffected.
\begin{algorithm}[H]
    \caption{Robust Adaptive Credal Loss (RACL)}
    \label{alg:rda}
    \begin{algorithmic}[1]
        \REQUIRE Training instance $(x, y) \in \cX \times \cY$, model prediction $\hatp(\vec{x}) \in \P(\cY)$, confidence threshold $\beta \in [0,1]$, Error rates for each class $e(c)$, Adjustment parameter $k$, $c$
        \STATE Adjust $\alpha(c)$ based on class-specific error rates $e(c)$ as specified in (6)
        \STATE Calculate $\pi$ as specified in (4)
        \STATE Calculate the credal set $Q_\pi$ derived from $\pi$ as specified in (7)
        \RETURN $\mathcal{L}^*(Q_\pi, \hat{p}(x))$ as specified in (9)
    \end{algorithmic}
\end{algorithm}

\section{Experiment}
\subsection{Datasets}
We conducted experiments on two fundus image datasets namely the OIA-ODIR dataset\cite{li2021benchmark} and the real-world dataset Kaggle Diabetic Retinopathy (DR) dataset\cite{kaggle_dr}. The OIA-ODIR dataset consists of 10,000 fundus images, and contains 8 classes of labels: normal, hypertensive retinopathy, glaucoma, cataract, age-related macular degeneration, hypertension complications, pathologic myopia, and other diseases/abnormalities. Noise was generated using our proposed clinical knowledge-based Asymmetric Noise Generation. The Kaggle DR dataset consists of 34,423 fundus photographs: one photograph per eye corresponding to the DR grading labels, and it is estimated that the initially published dataset has approximately 30$\%$ noisy labels in the initially released dataset [19]. In the experiment, the distribution of data across the training, validation, and test sets was configured in a ratio of 6:1:3. 
\subsection{Implementation Details}
All images were preprocessed to a size of 224x224. In the MAE pre-training phase, the mask ratio was set to 0.8, the batch size was set to 64, and the learning rate was set to $1 \times 10^{-4}$. Other hyperparameters of the MAE were set according to \cite{he2022masked}. For the fine-tuning phase, the learning rate was set to $1 \times 10^{-5}$, with a total of 30 training epochs. During the warmup phase, the learning rate was set to $1 \times 10^{-6}$, and the number of warmup epochs was set to 5. For RACL, we set $\beta_0 = 0.75$ and $\beta_1 = 0.55$. All experiments were conducted using PyTorch on an NVIDIA GeForce RTX 3090 GPU.

\subsection{Comparison with the State-of-the-Arts}
We compared our SSP-RACL framework with several state-of-the-art methods on the OIA-ODIR and Kaggle DR datasets, including Generalized Cross-Entropy (GCE) \cite{zhang2018generalized}, CORES \cite{cheng2020learning}, and Co-teaching \cite{RN728}. The performance metrics used for comparison were Average Precision (AP), Area Under the Curve (AUC), Precision, F1 score, and Recall. Detailed performance comparisons are shown in Table \ref{table:performance_comparison1} and Table \ref{table:performance_comparison2}, demonstrating that our method achieves the best performance across most metrics.
\begin{table}[htbp]
\centering
\caption{Performance comparison on the OIA-ODIR dataset}
\setlength{\tabcolsep}{13.2pt} % 调整列间距
\begin{tabular}{|c|c|c|c|c|}
\hline
Noise rate $r_n$ & \multicolumn{2}{c|}{\textbf{0.1}} & \multicolumn{2}{c|}{\textbf{0.2}} \\ \hline
& AP & AUC & AP & AUC \\ \hline
GCE\cite{zhang2018generalized} & 67.1 & 85.8 & 60.3 & 83.6 \\ \hline
CORES\cite{cheng2020learning} & 63.2 & 83.1 & 56.7 & 81.2 \\ \hline
Co-teaching\cite{RN728} & 71.5 & 85.5 & 69.6 & 83.1 \\ \hline
\textbf{Ours} & \textbf{72.4} & \textbf{86.4} & \textbf{71.9} & \textbf{84.3} \\ \hline
\end{tabular}
\label{table:performance_comparison1}
\end{table}

\begin{table}[htbp]
\centering
\caption{Performance comparison on the kaggle DR dataset}
\setlength{\tabcolsep}{14.5pt}
\begin{tabular}{|c|c|c|c|}
\hline
& Precision & F1 score & Recall \\ \hline
GCE\cite{zhang2018generalized} & 72 & 65.9 & 60.7 \\ \hline
CORES\cite{cheng2020learning} & 71.3 & 64.7 & 59.3 \\ \hline
Co-teaching\cite{RN728} & 80.4 & 77.9 & 75.6 \\ \hline
\textbf{Ours} & \textbf{82.7} & \textbf{80.6} & \textbf{78.6} \\ \hline
\end{tabular}
\label{table:performance_comparison2}
\end{table}
\subsection{Ablation Study}
To evaluate the contribution of each component in our SSP-RACL framework, we performed an ablation study by systematically removing different components and measuring the performance impact. The results, as shown in Table \ref{table:ablation_study}, demonstrate that both RACL and SSP significantly contribute to the overall performance of our model. 
\begin{table}[h]
    \centering
    \caption{Ablation study with $r_n=0.1$}
    \setlength{\tabcolsep}{33pt}
    \begin{tabular}{|c|c|c|}
        \hline
        SSP & RACL & AUC \\ \hline
        $\times$ & $\times$ & 81.7 \\ \hline
        $\checkmark$ & $\times$ & 84.3 \\ \hline
        $\times$ & $\checkmark$ & 85.2 \\ \hline
        $\checkmark$ & $\checkmark$ & 86.4 \\ \hline
    \end{tabular}
    \label{table:ablation_study}
\end{table}
\section*{Conclusion}
In our paper, we propose an SSP-RACL robust framework to address label noise in fundus image datasets. The clinical knowledge-based Asymmetric Noise Generation method simulates label noise as observed in clinical practice. The MAE captures the intrinsic structure of the data, reducing dependence on precise annotations. The RACL adjusts label relaxation parameters and focuses on hard-to-classify samples, effectively handling label noise. Experiments on OIA-ODIR and Kaggle DR datasets demonstrated significant performance improvements. Future work will further optimize these techniques and explore their application to other medical imaging tasks.
\section*{Acknowledgment}
This study was sponsored by Key Laboratory of Medical Electronics and Digital Health of Zhejiang Province (MEDC202301) and National Student Entrepreneurship Practice Program of Fudan University (202310246001S). 

\bibliographystyle{IEEEtran}
\bibliography{IEEEabrv,mybibfile}

% Generated by IEEEtran.bst, version: 1.12 (2007/01/11)
\begin{thebibliography}{10}
\providecommand{\url}[1]{#1}
\csname url@samestyle\endcsname
\providecommand{\newblock}{\relax}
\providecommand{\bibinfo}[2]{#2}
\providecommand{\BIBentrySTDinterwordspacing}{\spaceskip=0pt\relax}
\providecommand{\BIBentryALTinterwordstretchfactor}{4}
\providecommand{\BIBentryALTinterwordspacing}{\spaceskip=\fontdimen2\font plus
\BIBentryALTinterwordstretchfactor\fontdimen3\font minus \fontdimen4\font\relax}
\providecommand{\BIBforeignlanguage}[2]{{%
\expandafter\ifx\csname l@#1\endcsname\relax
\typeout{** WARNING: IEEEtran.bst: No hyphenation pattern has been}%
\typeout{** loaded for the language `#1'. Using the pattern for}%
\typeout{** the default language instead.}%
\else
\language=\csname l@#1\endcsname
\fi
#2}}
\providecommand{\BIBdecl}{\relax}
\BIBdecl

\bibitem{RN705}
O.~Ouda, E.~AbdelMaksoud, A.~Abd El-Aziz, and M.~Elmogy, ``Multiple ocular disease diagnosis using fundus images based on multi-label deep learning classification,'' \emph{Electronics}, vol.~11, no.~13, p. 1966, 2022.

\bibitem{RN706}
C.~Zhu, W.~Chen, T.~Peng, Y.~Wang, and M.~Jin, ``Hard sample aware noise robust learning for histopathology image classification,'' \emph{IEEE transactions on medical imaging}, vol.~41, no.~4, pp. 881--894, 2021.

\bibitem{RN728}
B.~Han, Q.~Yao, X.~Yu, G.~Niu, M.~Xu, W.~Hu, I.~Tsang, and M.~Sugiyama, ``Co-teaching: Robust training of deep neural networks with extremely noisy labels,'' \emph{Advances in neural information processing systems}, vol.~31, 2018.

\bibitem{zhou2021learning}
X.~Zhou, X.~Liu, C.~Wang, D.~Zhai, J.~Jiang, and X.~Ji, ``Learning with noisy labels via sparse regularization,'' in \emph{Proceedings of the IEEE/CVF international conference on computer vision}, 2021, pp. 72--81.

\bibitem{liu2020early}
S.~Liu, J.~Niles-Weed, N.~Razavian, and C.~Fernandez-Granda, ``Early-learning regularization prevents memorization of noisy labels,'' \emph{Advances in neural information processing systems}, vol.~33, pp. 20\,331--20\,342, 2020.

\bibitem{lu2021confidence}
Y.~Lu, Y.~Bo, and W.~He, ``Confidence adaptive regularization for deep learning with noisy labels,'' \emph{arXiv preprint arXiv:2108.08212}, 2021.

\bibitem{zhang2018generalized}
Z.~Zhang and M.~Sabuncu, ``Generalized cross entropy loss for training deep neural networks with noisy labels,'' \emph{Advances in neural information processing systems}, vol.~31, 2018.

\bibitem{cheng2020learning}
H.~Cheng, Z.~Zhu, X.~Li, Y.~Gong, X.~Sun, and Y.~Liu, ``Learning with instance-dependent label noise: A sample sieve approach,'' \emph{arXiv preprint arXiv:2010.02347}, 2020.

\bibitem{feng2021can}
L.~Feng, S.~Shu, Z.~Lin, F.~Lv, L.~Li, and B.~An, ``Can cross entropy loss be robust to label noise?'' in \emph{Proceedings of the twenty-ninth international conference on international joint conferences on artificial intelligence}, 2021, pp. 2206--2212.

\bibitem{xia2023combating}
X.~Xia, B.~Han, Y.~Zhan, J.~Yu, M.~Gong, C.~Gong, and T.~Liu, ``Combating noisy labels with sample selection by mining high-discrepancy examples,'' in \emph{Proceedings of the IEEE/CVF International Conference on Computer Vision}, 2023, pp. 1833--1843.

\bibitem{patel2023adaptive}
D.~Patel and P.~Sastry, ``Adaptive sample selection for robust learning under label noise,'' in \emph{Proceedings of the IEEE/CVF Winter Conference on Applications of Computer Vision}, 2023, pp. 3932--3942.

\bibitem{tanaka2018joint}
D.~Tanaka, D.~Ikami, T.~Yamasaki, and K.~Aizawa, ``Joint optimization framework for learning with noisy labels,'' in \emph{Proceedings of the IEEE conference on computer vision and pattern recognition}, 2018, pp. 5552--5560.

\bibitem{lu2022selc}
Y.~Lu and W.~He, ``Selc: self-ensemble label correction improves learning with noisy labels,'' \emph{arXiv preprint arXiv:2205.01156}, 2022.

\bibitem{wu2021learning}
Y.~Wu, J.~Shu, Q.~Xie, Q.~Zhao, and D.~Meng, ``Learning to purify noisy labels via meta soft label corrector,'' in \emph{Proceedings of the AAAI Conference on Artificial Intelligence}, vol.~35, no.~12, 2021, pp. 10\,388--10\,396.

\bibitem{cheng2021demystifying}
H.~Cheng, Z.~Zhu, X.~Sun, and Y.~Liu, ``Demystifying how self-supervised features improve training from noisy labels,'' 2021.

\bibitem{li2023disc}
Y.~Li, H.~Han, S.~Shan, and X.~Chen, ``Disc: Learning from noisy labels via dynamic instance-specific selection and correction,'' in \emph{Proceedings of the IEEE/CVF Conference on Computer Vision and Pattern Recognition}, 2023, pp. 24\,070--24\,079.

\bibitem{sheng2024adaptive}
M.~Sheng, Z.~Sun, Z.~Cai, T.~Chen, Y.~Zhou, and Y.~Yao, ``Adaptive integration of partial label learning and negative learning for enhanced noisy label learning,'' in \emph{Proceedings of the AAAI Conference on Artificial Intelligence}, vol.~38, no.~5, 2024, pp. 4820--4828.

\bibitem{albert2023your}
P.~Albert, E.~Arazo, T.~Krishna, N.~E. O’Connor, and K.~McGuinness, ``Is your noise correction noisy? pls: Robustness to label noise with two stage detection,'' in \emph{Proceedings of the IEEE/CVF Winter Conference on Applications of Computer Vision}, 2023, pp. 118--127.

\bibitem{ju2022improving}
L.~Ju, X.~Wang, L.~Wang, D.~Mahapatra, X.~Zhao, Q.~Zhou, T.~Liu, and Z.~Ge, ``Improving medical images classification with label noise using dual-uncertainty estimation,'' \emph{IEEE transactions on medical imaging}, vol.~41, no.~6, pp. 1533--1546, 2022.

\bibitem{hu2023fundus}
T.~Hu, B.~Yang, J.~Guo, W.~Zhang, H.~Liu, N.~Wang, and H.~Li, ``A fundus image classification framework for learning with noisy labels,'' \emph{Computerized Medical Imaging and Graphics}, vol. 108, p. 102278, 2023.

\bibitem{galdran2020cost}
A.~Galdran, J.~Dolz, H.~Chakor, H.~Lombaert, and I.~Ben~Ayed, ``Cost-sensitive regularization for diabetic retinopathy grading from eye fundus images,'' in \emph{Medical Image Computing and Computer Assisted Intervention--MICCAI 2020: 23rd International Conference, Lima, Peru, October 4--8, 2020, Proceedings, Part V 23}.\hskip 1em plus 0.5em minus 0.4em\relax Springer, 2020, pp. 665--674.

\bibitem{huang2023self}
S.-C. Huang, A.~Pareek, M.~Jensen, M.~P. Lungren, S.~Yeung, and A.~S. Chaudhari, ``Self-supervised learning for medical image classification: a systematic review and implementation guidelines,'' \emph{NPJ Digital Medicine}, vol.~6, no.~1, p.~74, 2023.

\bibitem{he2022masked}
K.~He, X.~Chen, S.~Xie, Y.~Li, P.~Doll{\'a}r, and R.~Girshick, ``Masked autoencoders are scalable vision learners,'' in \emph{Proceedings of the IEEE/CVF conference on computer vision and pattern recognition}, 2022, pp. 16\,000--16\,009.

\bibitem{lienen2023conformal}
J.~Lienen, C.~Demir, and E.~H{\"u}llermeier, ``Conformal credal self-supervised learning,'' in \emph{Conformal and Probabilistic Prediction with Applications}.\hskip 1em plus 0.5em minus 0.4em\relax PMLR, 2023, pp. 214--233.

\bibitem{wang2024creinnscredalsetintervalneural}
\BIBentryALTinterwordspacing
K.~Wang, K.~Shariatmadar, S.~K. Manchingal, F.~Cuzzolin, D.~Moens, and H.~Hallez, ``Creinns: Credal-set interval neural networks for uncertainty estimation in classification tasks,'' 2024. [Online]. Available: \url{https://arxiv.org/abs/2401.05043}
\BIBentrySTDinterwordspacing

\bibitem{lienen2024mitigating}
J.~Lienen and E.~H{\"u}llermeier, ``Mitigating label noise through data ambiguation,'' in \emph{Proceedings of the AAAI Conference on Artificial Intelligence}, vol.~38, no.~12, 2024, pp. 13\,799--13\,807.

\bibitem{caprio2024credallearningtheory}
\BIBentryALTinterwordspacing
M.~Caprio, M.~Sultana, E.~Elia, and F.~Cuzzolin, ``Credal learning theory,'' 2024. [Online]. Available: \url{https://arxiv.org/abs/2402.00957}
\BIBentrySTDinterwordspacing

\bibitem{dubois2001possibility}
D.~Dubois and H.~Prade, ``Possibility theory, probability theory and multiple-valued logics: A clarification,'' \emph{Annals of mathematics and Artificial Intelligence}, vol.~32, pp. 35--66, 2001.

\bibitem{hullermeier2014learning}
E.~H{\"u}llermeier, ``Learning from imprecise and fuzzy observations: Data disambiguation through generalized loss minimization,'' \emph{International Journal of Approximate Reasoning}, vol.~55, no.~7, pp. 1519--1534, 2014.

\bibitem{lienen2021label}
J.~Lienen and E.~H{\"u}llermeier, ``From label smoothing to label relaxation,'' in \emph{Proceedings of the AAAI conference on artificial intelligence}, vol.~35, no.~10, 2021, pp. 8583--8591.

\bibitem{li2021benchmark}
N.~Li, T.~Li, C.~Hu, K.~Wang, and H.~Kang, ``A benchmark of ocular disease intelligent recognition: One shot for multi-disease detection,'' in \emph{Benchmarking, Measuring, and Optimizing: Third BenchCouncil International Symposium, Bench 2020, Virtual Event, November 15--16, 2020, Revised Selected Papers 3}.\hskip 1em plus 0.5em minus 0.4em\relax Springer, 2021, pp. 177--193.

\bibitem{kaggle_dr}
\BIBentryALTinterwordspacing
Kaggle, ``Diabetic retinopathy detection,'' 2015, accessed: 2024-08-19. [Online]. Available: \url{https://www.kaggle.com/c/diabetic-retinopathy-detection/data}
\BIBentrySTDinterwordspacing

\end{thebibliography}

\end{document}